\documentclass[letterpaper, 10 pt, conference]{ieeeconf}  

\IEEEoverridecommandlockouts                              

\usepackage{graphics} 
\usepackage{epsfig} 
\usepackage{amsmath} 
\usepackage{amssymb}  
\usepackage{balance}
\usepackage{amsfonts}
\usepackage{booktabs} 
\usepackage{cite}
\usepackage{makecell}
\usepackage[backref]{hyperref}
\hypersetup{
hidelinks,
colorlinks=true,
linkcolor=black,
urlcolor=black,
citecolor=black
}
\usepackage{color}
\usepackage{soul}
\sethlcolor{yellow}
\graphicspath{{figures/}}
\usepackage{rotating}
\usepackage{pifont}
\usepackage{colortbl}
\usepackage{array} 
\usepackage{multirow}
\usepackage{graphicx}
\usepackage[table,xcdraw]{xcolor}

\usepackage{caption}
\newcommand{\cmark}{\ding{51}} 
\newcommand{\xmark}{\ding{55}} 

\definecolor{barriercolor}{RGB}{255, 120, 50}
\definecolor{bicyclecolor}{RGB}{255, 192, 203}
\definecolor{buscolor}{RGB}{255, 255, 0}
\definecolor{carcolor}{RGB}{0, 150, 245}
\definecolor{constructcolor}{RGB}{0, 255, 255}
\definecolor{motorcolor}{RGB}{200, 180, 0}
\definecolor{pedestriancolor}{RGB}{255, 0, 0}
\definecolor{trafficcolor}{RGB}{255, 240, 150}
\definecolor{trailercolor}{RGB}{135, 60, 0}
\definecolor{truckcolor}{RGB}{160, 32, 240}
\definecolor{drivablecolor}{RGB}{255, 0, 255}
\definecolor{otherflatcolor}{RGB}{139, 137, 137}
\definecolor{sidewalkcolor}{RGB}{75, 0, 75}
\definecolor{terraincolor}{RGB}{150, 240, 80}
\definecolor{manmadecolor}{RGB}{230, 230, 255}
\definecolor{vegetationcolor}{RGB}{0, 175, 0}
\definecolor{otherscolor}{RGB}{0, 0, 0}

\newcommand{\semkitfreq}[1]{\csname semkitfreq#1\endcsname}

\newcommand{\classfreq}[1]{{~\tiny(\semkitfreq{#1}\%)}}

\title{\LARGE \bf
SPOT-Occ: Sparse Prototype-guided Transformer for \\ Camera-based 3D Occupancy Prediction
}

\author{Suzeyu Chen$^{1}$, Leheng Li$^{1}$, and Ying-Cong Chen$^{1,2*}$
    \thanks{$^{1}$Artificial Intelligence Thrust, The Hong Kong University of Science and Technology (Guangzhou). 
    {\tt\footnotesize schen744@connect.hkust-gz.edu.cn,\newline lli181@connect.hkust-gz.edu.cn,\newline yingcongchen@hkust-gz.edu.cn}}
    \thanks{$^{2}$Department of Computer Science and Engineering, The Hong Kong University of Science and Technology. {\tt\footnotesize yingcongchen@ust.hk}}
    \thanks{$^{*}$Corresponding author}
}

\begin{document}

\maketitle
\thispagestyle{empty}
\pagestyle{empty}

\begin{abstract}
Achieving highly accurate and real-time 3D occupancy prediction from cameras is a critical requirement for the safe and practical deployment of autonomous vehicles. While this shift to sparse 3D representations solves the encoding bottleneck, it creates a new challenge for the decoder: how to efficiently aggregate information from a sparse, non-uniformly distributed set of voxel features without resorting to computationally prohibitive dense attention.
In this paper, we propose a novel Prototype-based Sparse Transformer Decoder that replaces this costly interaction with an efficient, two-stage process of guided feature selection and focused aggregation. Our core idea is to make the decoder's attention prototype-guided. We achieve this through a sparse prototype selection mechanism, where each query adaptively identifies a compact set of the most salient voxel features, termed prototypes, for focused feature aggregation.
To ensure this dynamic selection is stable and effective, we introduce a complementary denoising paradigm. This approach leverages ground-truth masks to provide explicit guidance, guaranteeing a consistent query-prototype association across decoder layers.
Our model, dubbed SPOT-Occ, outperforms previous methods with a significant margin in speed while also improving accuracy.
Source code is released at \url{https://github.com/chensuzeyu/SpotOcc}.
\end{abstract}

\section{Introduction}

Camera-based 3D semantic occupancy prediction offers a fine-grained 3D environmental understanding that surpasses traditional bounding boxes~\cite{huang2021bevdet, liu2023petrv2, liu2023sparsebev, wang2024towards, meng2024small, zhu2023curricular, shi2022srcn3d, pointpillar, bevdepth, zhou2023unidistill}, capable of representing complex geometric and semantic details, such as a car with a half-open door. Early methods constructed a dense 3D feature volume from images~\cite{cao2022monoscene, wei2023surroundocc, li2023fbocc, zhang2023occformer, wang2023openoccupancy, ma2023cotr, shi2024panossc, li2024hierarchical}, but the cubic computational complexity of this approach became a critical bottleneck for real-time applications. Recognizing that driving scenes are inherently sparse—for instance, approximately 67\% of voxels are empty in SemanticKITTI~\cite{behley2019semantickitti}—a paradigm shift towards sparse latent representations has emerged~\cite{li2023voxformer, tang2024sparseocc, liu2023fully, wang2024opus}. This shift effectively solved the encoding bottleneck but created a new, critical challenge for the decoder: how learnable queries can efficiently aggregate information from a sparse and unstructured set of voxel features.

\begin{figure}[t]
    \centering
    \includegraphics[width=8.6cm]{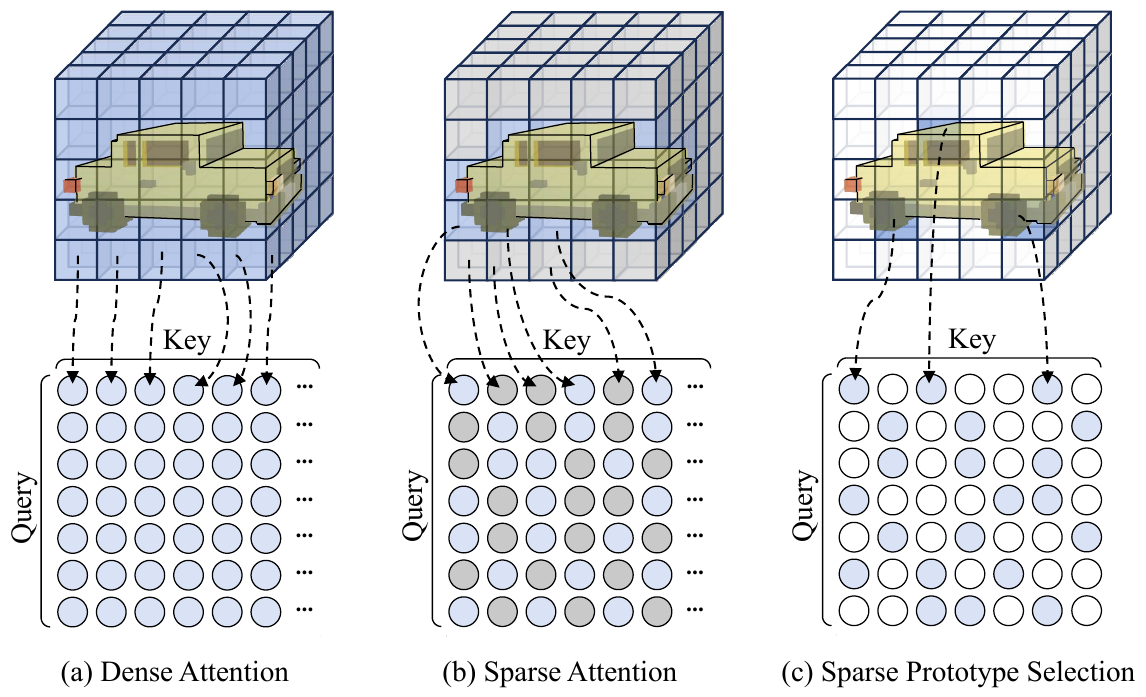}
    \caption{The matrix circles visualize query-key interactions; blue circles denote computed attention, while gray represent masked connections. 
        (a) Dense Attention interacts with all voxel features, leading to prohibitive cubic complexity. 
        (b) Sparse Attention masks empty voxels yet computes a full-sized attention matrix, limiting efficiency. 
        (c) Our Sparse Prototype Selection directly selects a compact set of salient features (prototypes) for each query, dramatically reducing complexity for efficient aggregation.}
    \vspace{-4pt}
    \label{fig:attention_complexity}
\end{figure}

This decoder-stage bottleneck is the central problem we address. As illustrated in Fig.~\ref{fig:attention_complexity}, the challenge has evolved with the development of prior methods. Early approaches (a) relied on dense attention~\cite{zhang2023occformer}, where interactions were computed between all queries and a dense voxel grid, leading to prohibitive computational complexity. With the shift to sparse representations, a seemingly straightforward solution (b) is to mask the attention for empty voxels~\cite{tang2024sparseocc, liu2023fully}. However, this approach remains inefficient as it still computes a full-sized attention matrix, merely setting some values to zero. This reveals a significant gap in designing decoders truly tailored for sparse 3D representations.

\begin{figure}[t]
    \centering
    \includegraphics[width=8.2cm]{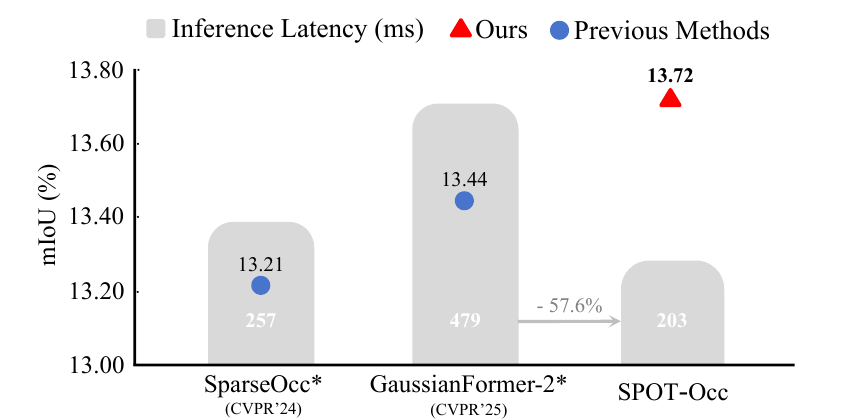}
    \caption{A quantitative benchmark of mIoU and latency on the nuScenes-Occupancy validation set. All latency are measured on a single NVIDIA RTX 3090 GPU with a batch size of 1, and * denotes results obtained by running the officially released code.}
    \label{fig:performance_compare}
\end{figure}

\begin{figure*}[t]
    \centering
    \includegraphics[width=17.2cm]{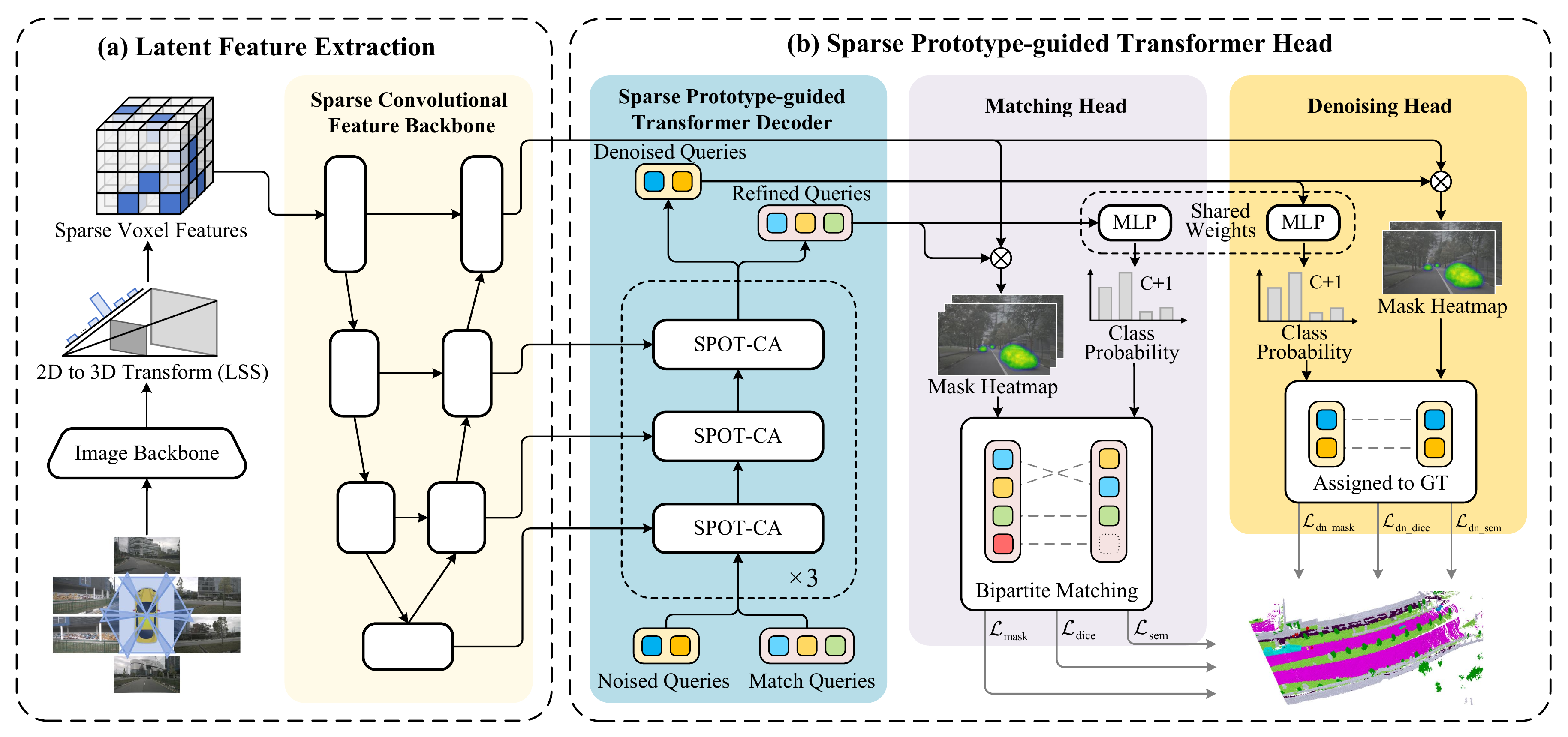}
    \vspace{-9pt}
    \caption{Overview of the proposed SPOT-Occ framework. (a) An image backbone~\cite{he2016deep} extracts multi-scale 2D features that are lifted to a sparse 3D space by LSS~\cite{philion2020lift} and refined with a sparse convolutional backbone~\cite{tang2024sparseocc}. (b) The decoder in our Sparse Prototype-guided Transformer Head refines queries via Sparse Prototype Selection under dual supervision. This includes a Denoising Head that leverages noised queries during training only, adding no overhead at inference time.}
    \vspace{-15pt}
    \label{fig:model}
\end{figure*}

In this work, we propose SPOT-Occ, a framework centered around a novel Prototype-based Sparse Transformer Decoder that resolves this challenge. Our core insight is to replace costly attention with an intelligent, two-stage process: guided feature selection followed by focused aggregation. Instead of interacting with all voxel features, each query learns to identify a compact, highly-relevant subset of features—its \textit{prototypes}—for attention, as shown in Fig.~\ref{fig:attention_complexity}(c). This is realized by our Sparse Prototype Selection strategy, which empowers queries to adaptively capture features from object-aware neighborhoods, dramatically reducing complexity.
To ensure this dynamic selection is not only efficient but also stable, we introduce a complementary denoising (DN) training paradigm. This ensures queries are well-initialized and robustly learn to engage with the correct sparse features. The main contributions of this paper are:
\begin{itemize}
    \item We propose SPOT-Occ, a novel sparse Transformer decoder that efficiently handles query-voxel interaction, circumventing the high cost of dense attention.
    \item We introduce a denoising training paradigm that provides stable supervision to guide our sparse decoder, enhancing prediction consistency without adding overhead at inference time.
    \item On the nuScenes-Occupancy benchmark, SPOT-Occ achieves higher accuracy while reducing inference latency by 57.6\% compared to GaussianFormer-2.
\end{itemize}

\begin{figure*}[t]
    \centering
    \includegraphics[width=17.4cm]{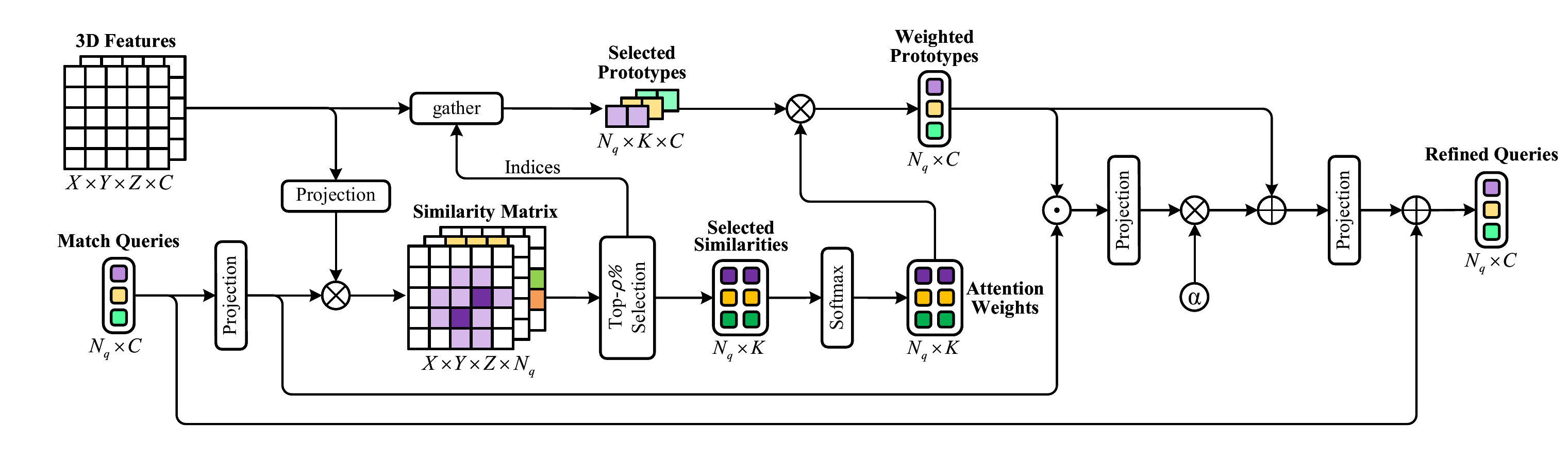}
    \vspace{-9pt}
    \caption{
        Cross-Attention of Sparse Prototype-guided Transformer Decoder (SPOT-CA). To efficiently process large-scale 3D features, our cross-attention mechanism introduces a sparse prototype selection step. Rather than comparing every query to every voxel feature, we first identify the Top-$\rho$\% most relevant features for each query. By creating sparse prototypes from only these key features, the decoder can refine the queries with significantly less computational cost while maintaining high performance.
    }
    \label{fig:spot_ca}
    \vspace{-13pt}
\end{figure*}

\section{Related Work}

\subsection{3D Occupancy Prediction}

The pursuit of an efficient and precise 3D scene representation has driven the development of occupancy prediction. Early methods using dense voxel grids~\cite{cao2022monoscene, wei2023surroundocc, li2023fbocc, zhang2023occformer, wang2023openoccupancy, ma2023cotr, shi2024panossc, li2024hierarchical} were computationally infeasible due to their cubic complexity. As a compromise, researchers turned to projected 2D representations like Bird's-Eye-View (BEV)~\cite{yu2023flashocc} and Tri-Perspective-View (TPV)~\cite{huang2023tri, cheng2021per}, sacrificing geometric fidelity for speed.
This trade-off became unnecessary with the introduction of sparse 3D representations. Inspired by point cloud processing, recent approaches~\cite{tang2024sparseocc, liu2023fully} process the 3D scene by operating only on the subset of relevant, non-empty voxels. This innovation effectively solved the encoding bottleneck but transferred the complexity to the decoder stage.

\subsection{Efficient Decoders for Transformer-based 3D Perception}

The paradigm of set prediction with Transformers, pioneered by DETR~\cite{carion2020end}, has reshaped modern perception by enabling end-to-end object detection. Its query-based design has inspired numerous advancements in both performance and efficiency~\cite{zhu2020deformable, meng2021conditional, yao2021efficient, sun2021sparse, liu2022dab, liu2023detrsm, jia2023detrhm}. This approach was successfully extended to segmentation by models such as MaskFormer~\cite{cheng2021per} and its successor Mask2Former~\cite{cheng2022masked}, which introduced masked attention to restrict computation to focused regions, proving highly effective for 2D tasks. The paradigm naturally found its way into the 3D domain for tasks like point cloud segmentation~\cite{schult23ICRA} and camera-based object detection~\cite{wang2022detr3d, liu2022petr, wang2023spetr, li2025bevformer}. However, the sheer scale of voxel features in 3D occupancy prediction, even within a sparse representation, renders these decoders inefficient. A naive masked attention~\cite{tang2024sparseocc, liu2023fully} fails to resolve this bottleneck, as it still computes a full-sized attention matrix.

\section{Methodology} 
\label{sec:method}

Our proposed SPOT-Occ framework, as illustrated in Fig.~\ref{fig:model}, is an end-to-end pipeline that begins with a latent feature extraction module. This module takes multi-view images as input, employing an image backbone with FPN~\cite{lin2017fpn} to extract 2D features. These features are then lifted into a 3D space via a Lift-Splat-Shoot (LSS)~\cite{philion2020lift} view transformer. The resulting sparse voxel features are further refined by a U-Net-like sparse convolutional backbone~\cite{tang2024sparseocc} to produce a multi-scale 3D feature pyramid. Next, these sparse 3D features are processed by our novel Sparse Prototype-guided Transformer Decoder (Sec.~\ref{sub:sparse_proto_select}). Following Mask2Former~\cite{cheng2022masked}, each refined query is passed to prediction heads, generating a class probability and an output vector that produces a mask heatmap via a dot product with the voxel features. The final semantic prediction for each voxel is obtained by taking the argmax across all queries on their class-scaled mask heatmaps. For training, these outputs are supervised in parallel by a Matching Head and a Denoising Head (Sec.~\ref{sub:mask_piloted_training}). The entire framework is trained end-to-end under the supervision of a composite objective function, which is detailed in Sec.~\ref{sub:loss}.

\subsection{Sparse Prototype Selection}
\label{sub:sparse_proto_select}

The central challenge in decoding sparse voxel features lies in efficiently connecting learnable object queries to a vast and non-uniformly distributed set of features. A naive dense cross-attention, with a complexity of $\mathcal{O}(N_q \cdot N_v \cdot C)$, where $N_v = X \times Y \times Z$ denotes the total number of voxels in the 3D feature grid, would form a computational bottleneck. To overcome this, we introduce a query-guided prototype selection mechanism, which re-frames the interaction into an efficient two-stage process: dynamic feature selection followed by focused aggregation.

\textbf{Deformable Top-$\rho$\% Selection.}
Our core intuition is that instead of performing costly attention over all sparse voxel features, a query can more efficiently gather information by focusing on a compact subset of the most salient ones. We term these dynamically selected features \textit{prototypes}, as they serve as the most representative exemplars for understanding a specific object or region. This is operationalized through Deformable Top-$\rho$\% Selection mechanism. The term \textit{deformable} signifies the adaptive selection of prototypes from multi-scale sparse features, enabling the receptive field to conform to the object's geometric structure.

The selection process begins by computing a saliency score for each query-key pair, defined as the cosine similarity between their projected vectors:
\begin{equation} S(\boldsymbol{q}, \boldsymbol{k}) = \frac{\text{Proj}_Q(\boldsymbol{q}) \cdot \text{Proj}_K(\boldsymbol{k})}{\|\text{Proj}_Q(\boldsymbol{q})\| \|\text{Proj}_K(\boldsymbol{k})\|}, \end{equation}
where $\boldsymbol{q} \in \mathbb{R}^C$ and $\boldsymbol{k} \in \mathbb{R}^C$ are query and key vectors, and $\text{Proj}_Q, \text{Proj}_K$ are linear layers. For each query, we select the Top-$\rho$\% keys with the highest scores to form a prototype set of size $k = \lceil \rho \cdot N_v \rceil$, where $N_v$ is the number of active voxels.

In our multi-head implementation, this selection process is executed in parallel across all attention heads. The query, key, and value tensors are first projected and split into $H$ heads along the feature dimension. The saliency score computation and the subsequent Top-$\rho$\% selection are performed independently for each head. This allows each head to learn to focus on a distinct, specialized subset of prototype features based on different representational subspaces.

\textbf{Prototype-guided Aggregation and Query Refinement.}
Once the prototypes are identified, they are aggregated into a single feature vector. First, attention weights are computed by applying a temperature-scaled softmax to the saliency scores of the selected prototypes:
\begin{equation} \boldsymbol{\alpha} = \text{softmax}(\boldsymbol{S}_{\mathcal{I}} / \sqrt{C}), \end{equation}
where $\boldsymbol{\alpha} \in \mathbb{R}^k$ is the vector of attention weights, $\boldsymbol{S}_{\mathcal{I}}$ is the vector of saliency scores for the selected prototypes (indexed by $\mathcal{I}$), and $\sqrt{C}$ is a standard scaling factor.
These weights guide the aggregation of the corresponding prototype values into a single feature vector $\boldsymbol{v}^{\text{agg}}$:
\begin{equation} \boldsymbol{v}^{\text{agg}} = \sum_{j=1}^{k} \alpha_j \boldsymbol{v}_j, \end{equation}
This vector $\boldsymbol{v}^{\text{agg}}$ represents the weighted sum of the prototype values $\{\boldsymbol{v}_j\}_{j=1}^k$, where each prototype is modulated by its attention weight $\alpha_j$.
Finally, the original query $\boldsymbol{q}$ is updated by integrating this aggregated information through a gating mechanism. First, an interaction vector $\boldsymbol{i}$ is computed:
\begin{equation} \boldsymbol{i} = \text{FFN}_1(\text{Proj}_Q(\boldsymbol{q}) \odot \boldsymbol{v}^{\text{agg}}), \end{equation}
where $\odot$ denotes element-wise multiplication. This vector is then processed to produce an output gate $\boldsymbol{o}$:
\begin{equation} \boldsymbol{o} = \text{FFN}_2(\alpha \cdot \text{Norm}(\boldsymbol{i}) + \boldsymbol{v}^{\text{agg}}), \end{equation}
In this step, $\text{Norm}$ denotes Layer Normalization and $\alpha$ is a learnable scalar. The query is finally updated via a residual connection:
\begin{equation} \boldsymbol{q}^{\text{refined}} = \boldsymbol{q} + \text{Dropout}(\boldsymbol{o}), \end{equation}
This update formulation, featuring a gated mechanism, offers a more sophisticated feature integration than standard attention, thereby enhancing the stability of the training process.
This entire approach reduces the complexity of cross-attention from $\mathcal{O}(N_q N_v)$ to just $\mathcal{O}(N_q k)$, where $k \ll N_v$. By constraining each query to interact with only a small fraction of the most relevant non-empty features, we achieve significant computational savings while empowering each query to dynamically construct an object-aware receptive field from the sparse 3D feature space.

\begin{figure}[t]
    \centering
    \includegraphics[width=1.0\linewidth]{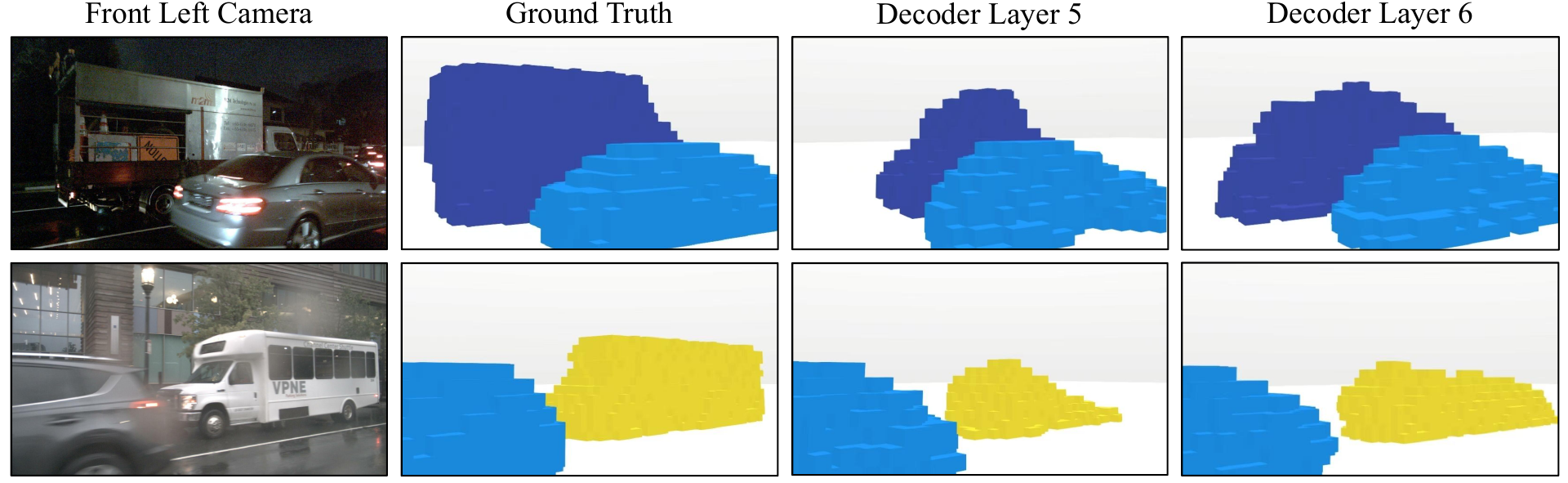}
    \caption{This figure illustrates how the same query predicts significantly inconsistent masks across consecutive decoder layers (Layer 5 vs. Layer 6). This observed instability motivates our denoising training strategy to ensure stable optimization.}
    \label{fig:layer_dif_vis}
\end{figure}

\begin{table}[t]
    \centering
    \caption{The table reports the Layer-wise Mean IoU (\%). Our Denoising training on SPOT (Sparse Prototype-guided Transformer) significantly stabilizes predictions across decoder layers.}
    \label{tab:layer_consistency}
    \resizebox{\linewidth}{!}{%
    \begin{tabular}{lccccccccc}
    \toprule
    Decoder Layer Number & 1 & 2 & 3 & 4 & 5 & 6 & 7 & 8 & 9 \\
    \midrule
    Sparse Mask2Former~\cite{tang2024sparseocc} & 21.2 & 45.1 & 53.6 & 62.5 & 63.1 & 62.8 & 70.2 & 68.9 & 68.1 \\
    SPOT w/o DN Training & 23.5 & 54.9 & 58.1 & 65.4 & 64.9 & 63.5 & 71.7 & 70.5 & 69.8 \\
    SPOT w/ DN Training & \textbf{48.8} & \textbf{81.7} & \textbf{83.5} & \textbf{91.6} & \textbf{91.3} & \textbf{89.5} & \textbf{93.1} & \textbf{92.9} & \textbf{92.3} \\
    \bottomrule
    \end{tabular}%
    }
\end{table}

\subsection{Denoising Training for Stable Selection}
\label{sub:mask_piloted_training}
While our Deformable Top-$\rho$\% selection mechanism grants queries the flexibility to adapt to object geometry, this dynamic process can introduce instability during training. Specifically, as a query is refined across different decoder layers, it interacts with voxel prototypes of varying scales.
As visualized in Fig.~\ref{fig:layer_dif_vis},  a single query to predict significantly inconsistent masks between layers, which in turn impairs training stability.

To systematically quantify this phenomenon, we introduce the Layer-wise Mean IoU. This metric measures the average Intersection over Union (IoU) of masks predicted by the same query across adjacent decoder layers. For a given layer $l$, it is formally defined as:
\vspace{-2pt}
\begin{equation}
\text{LM-IoU}^{(l)} = \frac{1}{N_q} \sum_{q=1}^{N_q} \frac{|M_{q}^{(l-1)} \cap M_{q}^{(l)}|}{|M_{q}^{(l-1)} \cup M_{q}^{(l)}|},
\end{equation}
\vspace{-2pt}
where $M_{q}^{(l-1)}$ and $M_{q}^{(l)}$ are the masks predicted by the $q$-th query at layers $l-1$ and $l$ respectively, and $N_q$ is the total number of queries. As shown in Table~\ref{tab:layer_consistency}, both the baseline model and our SPOT without denoising exhibit low consistency, especially in early layers.

Inspired by denoising methods~\cite{li2024dn}, we introduce a denoising training strategy to provide more direct and stable supervision. During training, we augment the standard learnable \textit{match queries} with a set of \textit{noised queries}. Specifically, for each ground-truth (GT) object, we generate a corresponding query from its class embedding, which is then intentionally corrupted.

This corruption involves two types of noise: \textit{semantic noise}, introduced by randomly flipping the GT class label, and \textit{feature noise}, where a random perturbation is added to the query's feature embedding. The feature noising process is formulated as:
\begin{equation}
\boldsymbol{q}'_{\text{feat}} = \boldsymbol{e}_{\text{gt}} + \boldsymbol{\delta},
\end{equation}
where $\boldsymbol{e}_{\text{gt}}$ is the GT class embedding and $\boldsymbol{\delta}$ is a noise vector sampled from a Gaussian distribution.

As illustrated in Fig.~\ref{fig:model}, the noised queries and match queries are concatenated and processed together by the decoder layers. However, their prototype selection is guided differently: noised queries are guided by the GT masks, forcing them to focus on correct spatial regions from the outset, whereas match queries use their own predicted masks for guidance. After decoder refinement, the queries are directed to two separate heads. The noised queries are tasked with reconstructing the original GT class and mask, providing a stable, one-to-one supervision signal that bypasses the Hungarian matching used for match queries. Notably, both the Denoising Head and the Matching Head employ a shared MLP for class prediction. This entire denoising framework is active only during training, adding no computational overhead at inference time.

\begin{table*}[t]
    \scriptsize
    \setlength{\tabcolsep}{0.0035\linewidth}
    \centering
    \caption{Quantitative results on the nuScenes-Occupancy~\cite{wang2023openoccupancy} validation set. We report geometric accuracy (IoU), semantic quality (mIoU), a detailed per-class IoU breakdown, and latency (ms), benchmarked on a single NVIDIA RTX 3090 (batch=1). An asterisk (*) denotes results from official code. All methods use 192 channels; GaussianFormer-2~\cite{huang2024gaussianformer-2} is configured with 12,800 Gaussians. The C is camera input, and bold marks the best performance.}
    \resizebox{1\linewidth}{!}
    {
    \begin{tabular}{l|c| c c | c c c c c c c c c c c c c c c c | c}

        \hline
        Method
        & \makecell[c]{Input}
        & \rotatebox{90}{\textbf{IoU} $\uparrow$}
        & \rotatebox{90}{\textbf{mIoU} $\uparrow$}
        & \rotatebox{90}{\textcolor{barriercolor}{$\blacksquare$} \textbf{barrier} $\uparrow$} 
        & \rotatebox{90}{\textcolor{bicyclecolor}{$\blacksquare$} \textbf{bicycle} $\uparrow$}
        & \rotatebox{90}{\textcolor{buscolor}{$\blacksquare$} \textbf{bus} $\uparrow$} 
        & \rotatebox{90}{\textcolor{carcolor}{$\blacksquare$} \textbf{car} $\uparrow$} 
        & \rotatebox{90}{\textcolor{constructcolor}{$\blacksquare$} \textbf{const. veh.} $\uparrow$} 
        & \rotatebox{90}{\textcolor{motorcolor}{$\blacksquare$} \textbf{motorcycle} $\uparrow$} 
        & \rotatebox{90}{\textcolor{pedestriancolor}{$\blacksquare$} \textbf{pedestrian} $\uparrow$} 
        & \rotatebox{90}{\textcolor{trafficcolor}{$\blacksquare$} \textbf{traffic cone} $\uparrow$} 
        & \rotatebox{90}{\textcolor{trailercolor}{$\blacksquare$} \textbf{trailer} $\uparrow$} 
        & \rotatebox{90}{\textcolor{truckcolor}{$\blacksquare$} \textbf{truck} $\uparrow$} 
        & \rotatebox{90}{\textcolor{drivablecolor}{$\blacksquare$} \textbf{drive. suf.} $\uparrow$} 
        & \rotatebox{90}{\textcolor{otherflatcolor}{$\blacksquare$} \textbf{other flat} $\uparrow$} 
        & \rotatebox{90}{\textcolor{sidewalkcolor}{$\blacksquare$} \textbf{sidewalk} $\uparrow$} 
        & \rotatebox{90}{\textcolor{terraincolor}{$\blacksquare$} \textbf{terrain} $\uparrow$} 
        & \rotatebox{90}{\textcolor{manmadecolor}{$\blacksquare$} \textbf{manmade} $\uparrow$} 
        & \rotatebox{90}{\textcolor{vegetationcolor}{$\blacksquare$} \textbf{vegetation} $\uparrow$}
        & \makecell[c]{3D / Overall \\Latency (ms) $\downarrow$}
        \\
        \hline\hline
        MonoScene~\cite{cao2022monoscene} & C & 18.4 & 6.9 & 7.1  & 3.9  &  9.3 &  7.2 & 5.6  & 3.0  &  5.9& 4.4& 4.9 & 4.2 & 14.9 & 6.3  & 7.9 & 7.4 & \textbf{10.0} & 7.6 & - \\
        TPVFormer~\cite{huang2023tri} &C & 15.3 & 7.8 & 9.3 & 4.1  &  11.3 &  10.1 & 5.2  & 4.3  & 5.9 & 5.3&  6.8& 6.5 & 13.6 & 9.0  & 8.3 & 8.0  & 9.2 & 8.2 & - \\
        OpenOccupancy~\cite{wang2023openoccupancy}  & C & 19.3 & 10.3  &  9.9 & 6.8  & 11.2  & 11.5  & 6.3  & 8.4  & 8.6 & 4.3 & 4.2 & 9.9 & 22.0  & 15.8 & 14.1  & 13.5  & 7.3 & 10.2 & - \\
        C-CONet~\cite{wang2023openoccupancy} & C & 20.1 & 12.8 &13.2  & 8.1 & 15.4 &  17.2 & 6.3  & \textbf{11.2}  & 10.0  &  8.3 & 4.7 & 12.1 & 31.4 & 18.8 & 18.7 & 16.3 & 4.8 & 8.2 & - \\
        SparseOcc*~\cite{tang2024sparseocc} & C & 20.3 & 13.2 & 16.1 & 7.3 & 15.4 & 17.6 & 5.9 & 9.2 & 10.5 & 9.5 & 5.8 & 12.8 & 30.7 & 21.1 & 19.0 & \textbf{16.5} & 5.0 & 8.7 & 195 / 257 \\
        GaussianFormer-2*~\cite{huang2024gaussianformer-2} & C & \textbf{20.6} & 13.4 & 15.9 & 7.2 & \textbf{18.1} & \textbf{19.3} & 6.7 & 10.5 & 8.9 & 6.7 & \textbf{9.1} & \textbf{14.5} & 26.4 & 15.4 & 17.3 & 15.3 & 8.7 & \textbf{14.4} & 352 / 479 \\
        \hline
        \rowcolor{gray!30}
        SPOT-Occ (ours) & C & 20.5 & \textbf{13.7} & \textbf{16.4} & \textbf{7.6} & 16.5 & 18.0 & \textbf{6.9} & 9.3 & \textbf{11.1} & \textbf{9.9} & 6.5 & 12.9 & \textbf{31.7} & \textbf{22.1} & \textbf{19.4} & 16.1 & 5.4 & 9.7 & \textbf{141 / 203} \\
        \hline
    \end{tabular}}
    \vspace{-8pt}
    \label{tab:nusc}
\end{table*}

\begin{table*}[t]
    \scriptsize
    \setlength{\tabcolsep}{0.0035\linewidth}
    \centering
    \caption{Semantic scene completion performance on the SemanticKITTI~\cite{behley2019semantickitti} validation set. We evaluate using geometric IoU, semantic mIoU, and a per-class breakdown. An asterisk (*) marks RGB-input variants from~\cite{cao2022monoscene} for fair comparison. A dagger (†) denotes results obtained from official public code. The C signifies camera input, and bold marks the best performance.}
    \resizebox{1\linewidth}{!}{
    \begin{tabular}{l|c|c c| c c c c c c c c c c c c c c c c c c c}
        \hline
        Method & Input & \textbf{IoU} $\uparrow$ & \textbf{mIoU} $\uparrow$
        & \rotatebox{90}{\textbf{road}\classfreq{road} $\uparrow$} 
        & \rotatebox{90}{\textbf{sidewalk}\classfreq{sidewalk} $\uparrow$}
        & \rotatebox{90}{\textbf{parking}\classfreq{parking} $\uparrow$} 
        & \rotatebox{90}{\textbf{other-ground}\classfreq{otherground} $\uparrow$} 
        & \rotatebox{90}{\textbf{building}\classfreq{building} $\uparrow$} 
        & \rotatebox{90}{\textbf{car}\classfreq{car} $\uparrow$} 
        & \rotatebox{90}{\textbf{truck}\classfreq{truck} $\uparrow$} 
        & \rotatebox{90}{\textbf{bicycle}\classfreq{bicycle} $\uparrow$} 
        & \rotatebox{90}{\textbf{motorcycle}\classfreq{motorcycle} $\uparrow$} 
        & \rotatebox{90}{\textbf{other-vehicle}\classfreq{othervehicle} $\uparrow$} 
        & \rotatebox{90}{\textbf{vegetation}\classfreq{vegetation} $\uparrow$} 
        & \rotatebox{90}{\textbf{trunk}\classfreq{trunk} $\uparrow$} 
        & \rotatebox{90}{\textbf{terrain}\classfreq{terrain} $\uparrow$} 
        & \rotatebox{90}{\textbf{person}\classfreq{person} $\uparrow$} 
        & \rotatebox{90}{\textbf{bicyclist}\classfreq{bicyclist} $\uparrow$} 
        & \rotatebox{90}{\textbf{motorcyclist}\classfreq{motorcyclist} $\uparrow$} 
        & \rotatebox{90}{\textbf{fence}\classfreq{fence} $\uparrow$} 
        & \rotatebox{90}{\textbf{pole}\classfreq{pole} $\uparrow$} 
        & \rotatebox{90}{\textbf{traffic-sign}\classfreq{trafficsign} $\uparrow$} 
        \\
        \hline\hline
        LMSCNet*~\cite{lmscnet} & C & 28.61 & 6.70 & 40.68 & 18.22 & 4.38 & 0.00 & 10.31 & 18.33 & 0.00 & 0.00 & 0.00 & 0.00 & 13.66 & 0.02 & 20.54 & 0.00 & 0.00 & 0.00 & 1.21 & 0.00 & 0.00  \\ %
        3DSketch*~\cite{3d-sketch} & C & 33.30 & 7.50 & 41.32 & 21.63 & 0.00 & 0.00 & 14.81 & 18.59 & 0.00 & 0.00 & 0.00 & 0.00 & \textbf{19.09} & 0.00 & 26.40 & 0.00 & 0.00 & 0.00 & 0.73 & 0.00 & 0.00 \\ %
        AICNet*~\cite{li2020anisotropic} & C & 29.59 & 8.31 & 43.55 & 20.55 & 11.97 & 0.07 & 12.94 & 14.71 & 4.53 & 0.00 & 0.00 & 0.00 & 15.37 & 2.90 & 28.71 & 0.00 & 0.00 & 0.00 & 2.52 & 0.06 & 0.00  \\ %
        JS3C-Net*~\cite{js3cnet} & C & \textbf{38.98} & 10.31 & 50.49 & 23.74 & 11.94 & 0.07 & 15.03 & 24.65 & 4.41 & 0.00 & 0.00 & 6.15 & 18.11 & \textbf{4.33} & 26.86 & 0.67 & 0.27 & 0.00 & 3.94 & 3.77 & 1.45 \\
        MonoScene~\cite{cao2022monoscene} & C & 36.86 & 11.08 & 56.52 & 26.72 & 14.27 & 0.46 & 14.09 & 23.26 & 6.98 & 0.61 & 0.45 & 1.48 & 17.89 & 2.81 & 29.64 & 1.86 & 1.20 & 0.00 & 5.84 & 4.14 & 2.25 \\
        TPVFormer~\cite{huang2023tri} & C & 35.61 & 11.36 & 56.50 & 25.87 & 20.60 & 0.85 & 13.88 & 23.81 & 8.08 & 0.36 & 0.05 & 4.35 & 16.92 & 2.26 & 30.38 & 0.51 & 0.89 & 0.00 & 5.94 & 3.14 & 1.52 \\
        SparseOcc$\dagger$~\cite{tang2024sparseocc} & C & 35.92 & 12.21 & \textbf{59.67} & 28.06 & 13.23 & 0.14 & \textbf{15.33} & 25.12 & 9.22 & 0.68 & 1.49 & 8.64 & 19.01 & 3.51 & 30.93 & \textbf{2.29} & \textbf{1.21} & 0.00 & 6.59 & \textbf{4.24} & \textbf{2.67} \\
        \hline
        \rowcolor{gray!30}
        SPOT-Occ (ours) & C & 36.40 & \textbf{13.27} & 59.33 & \textbf{28.46} & \textbf{21.04} & \textbf{1.26} & 15.19 & \textbf{26.42} & \textbf{16.10} & \textbf{1.08} & \textbf{1.59} & \textbf{11.88} & 18.97 & 3.03 & \textbf{31.57} & 1.82 & 1.16 & 0.00 & \textbf{6.61} & 4.10 & 2.56 \\
        \hline
    \end{tabular}}
    \vspace{-8pt}
    \label{table:kitti}
    \vspace{-10pt}
\end{table*}

\subsection{Objective Function}
\label{sub:loss}
Our model is trained with a composite loss function, combining losses for the Transformer decoder and the view transformation module. The decoder is supervised at each layer via a matching loss $\mathcal{L}_{\text{match}}$ and a denoising loss $\mathcal{L}_{\text{dn}}$, while an auxiliary depth loss $\mathcal{L}_{\text{depth}}$ supervises the view transformer. The overall objective is:
\begin{equation}
\mathcal{L} = \mathcal{L}_{\text{match}} + \mathcal{L}_{\text{dn}} + \mathcal{L}_{\text{depth}},
\end{equation}
Both $\mathcal{L}_{\text{match}}$ and $\mathcal{L}_{\text{dn}}$ are a weighted sum of cross-entropy loss for classification and a combination of Binary Cross-Entropy and Dice losses for mask prediction. While $\mathcal{L}_{\text{match}}$ uses the Hungarian algorithm to find an optimal assignment to ground-truth objects, $\mathcal{L}_{\text{dn}}$ leverages a fixed, one-to-one correspondence. The depth loss provides direct supervision to the LSS module using ground-truth depth from LiDAR. For efficiency, mask losses are calculated on a sampled subset of 3D points.

\section{Experiment}
\subsection{Datasets}

Our primary evaluation is conducted on the nuScenes-Occupancy~\cite{wang2023openoccupancy} benchmark, which is derived from the large-scale nuScenes dataset and features dense 3D semantic occupancy annotations. We adhere to the official data split, utilizing 700 scenes for training and 150 for validation. The ground truth encompasses 17 categories (16 semantic plus one empty class) within a spatial volume of $-40$m to $40$m along the X/Y axes and $-1$m to $5.4$m along the Z axis.

We also report results on the SemanticKITTI~\cite{behley2019semantickitti} dataset for the monocular semantic scene completion task. Following the standard protocol, we train on sequences 00-10 (excluding 08) and validate on sequence 08. This dataset provides annotations for 19 distinct semantic classes alongside an empty class for each 3D voxel.

\begin{figure*}[t]
    \centering
    \includegraphics[width=17.7cm]{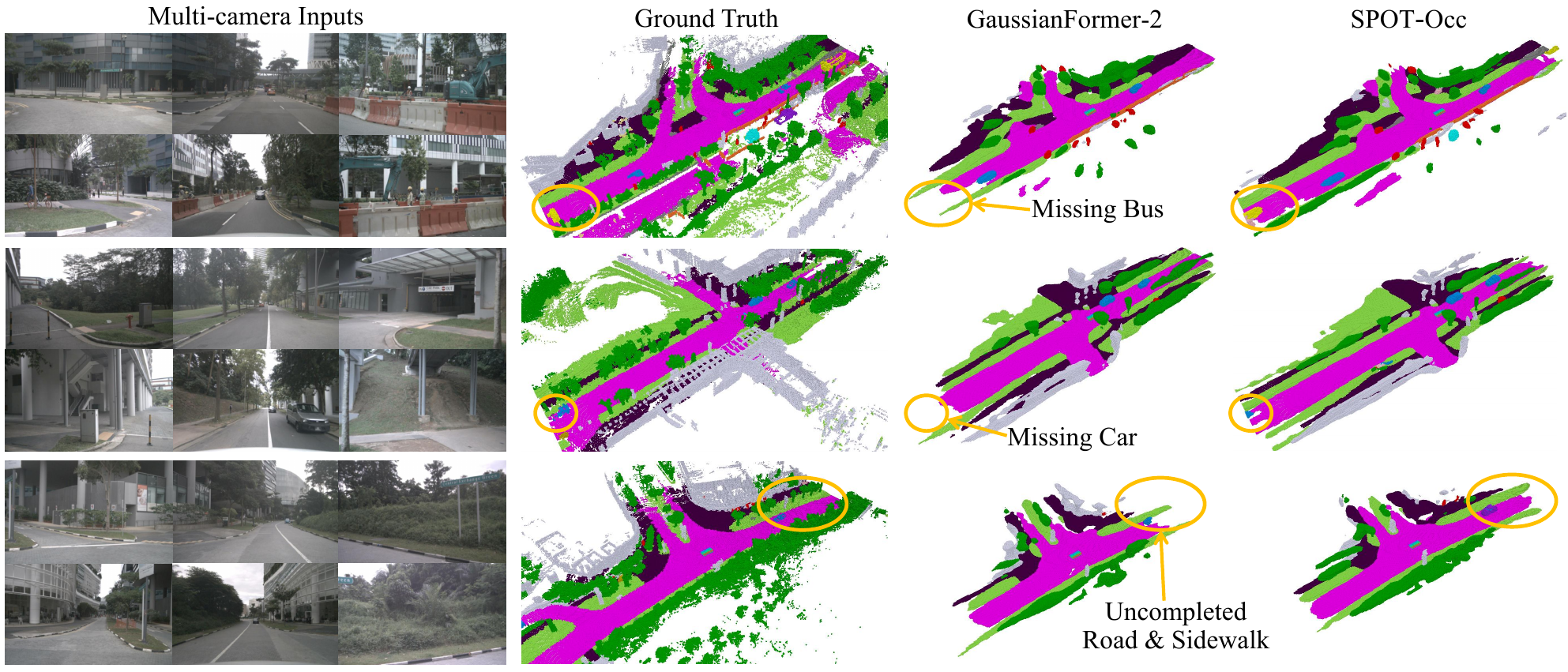}
    \vspace{-18pt}
    \caption{Qualitative comparison of our method and GaussianFormer-2 \cite{huang2024gaussianformer-2} on the nuScenes-Occuppancy dataset.}
    \vspace{-16pt}
    \label{fig:vis}
\end{figure*}

\subsection{Implementation Details}

We employ ResNet-50~\cite{he2016deep} for nuScenes-Occupancy and EfficientNetB7~\cite{tan2019efficientnet} for SemanticKITTI. The view transformation is realized via the Lift-Splat-Shoot (LSS)~\cite{philion2020lift} paradigm. Our core contribution, the Sparse Prototype-guided Transformer Decoder, is constructed with 9 layers and processes $N_q=100$ learnable queries. A key hyperparameter, the prototype sampling ratio, is set to $\rho=0.08$. We train the model end-to-end with the AdamW optimizer~\cite{loshchilov2017decoupled}, setting the learning rate to $2 \times 10^{-4}$ and weight decay to 0.01. The model is trained for 24 epochs on nuScenes and 30 on SemanticKITTI with a batch size of 8 on 8 NVIDIA RTX 4080 GPUs. During inference, the denoising branch is disabled and final masks are upsampled via trilinear interpolation (4$\times$ for nuScenes, 2$\times$ for SemanticKITTI) to match the ground-truth resolution.

\subsection{Comparison with the State-of-the-Art}

We benchmark SPOT-Occ against leading methods on the nuScenes-Occupancy and SemanticKITTI datasets. As shown in Table~\ref{tab:nusc} and Table~\ref{table:kitti}, our approach achieves a compelling balance of high accuracy and computational efficiency.

\textbf{Quantitative Analysis.} 
On the nuScenes-Occupancy validation set (Table~\ref{tab:nusc}), SPOT-Occ achieves  the highest mIoU of 13.7\%, surpassing prior works such as SparseOcc (13.2\%) and GaussianFormer-2 (13.4\%). Notably, this performance is attained with a 57.6\% reduction in inference latency compared to the latter, underscoring the efficacy of our sparse prototype-guided decoder in mitigating the computational burden of traditional attention mechanisms. This competitive performance is also reflected on the SemanticKITTI benchmark (Table~\ref{table:kitti}), where our model secures the highest mIoU of 13.27\%.

\textbf{Qualitative Analysis.} 
As illustrated in Fig.~\ref{fig:vis}, our method yields predictions with higher geometric fidelity and semantic precision than strong baselines. SPOT-Occ better reconstructs fine-grained details, such as recovering \textit{vehicles} missed by the baseline and completing the geometry of the \textit{road} and \textit{sidewalk}. These improvements validate our prototype-guided attention's ability to capture object-aware features for a more faithful 3D scene completion.

\begin{table}[ht]
    \centering
    \caption{Analysis of Key Components. We validate the effectiveness of the Sparse Prototype-guided Cross-Attention (SPOT-CA) and the Denoising (DN) training paradigm. The baseline is SparseOcc~\cite{tang2024sparseocc}.}
    \label{tab:ablation_main}
    \resizebox{0.9\linewidth}{!}{%
    \begin{tabular}{cc|cc}
    \toprule
    \multicolumn{2}{c|}{Components} & \multicolumn{2}{c}{Metrics} \\
    \cmidrule(r){1-2} \cmidrule(l){3-4}
    SPOT-CA & DN Training & mIoU (\%) $\uparrow$ & Overall Latency (ms) $\downarrow$ \\
    \midrule
    \xmark & \xmark & 12.21 & 211 \\
    \cmark & \xmark & 12.85 & 166 \\
    \xmark & \cmark & 12.59 & 209 \\
    \rowcolor{gray!20}
    \cmark & \cmark & \textbf{13.27} & \textbf{164} \\
    \bottomrule
    \end{tabular}%
    }
\end{table}

\subsection{Ablation Study}
\label{sec:ablation}

To validate the effectiveness of our proposed SPOT-Occ, we conduct a series of ablation experiments on the SemanticKITTI validation set. We systematically analyze our two core contributions: the Sparse Prototype-guided Cross-Attention (SPOT-CA) and the Denoising (DN) training paradigm, along with their key design choices.

\textbf{Effectiveness of Key Components.}
As shown in Table~\ref{tab:ablation_main}, our two core designs bring significant improvements over the baseline. (1) Integrating SPOT-CA boosts the mIoU by 0.64\% (to 12.85\%) while concurrently cutting latency by 21.3\%. This confirms it captures salient features more efficiently. (2) The Denoising (DN) training paradigm alone improves mIoU by 0.38\% without any inference overhead. (3) When combined, our full SPOT-Occ model achieves the best performance (13.27\% mIoU), demonstrating the synergy between our proposed components.

\textbf{Analysis of Design Choices.}
We further analyze the key hyperparameters. For the prototype ratio in SPOT-CA, Table~\ref{tab:ablation_rho} shows that $\rho=0.08$ strikes the optimal balance between feature richness and computational cost. For the DN training, as detailed in Table~\ref{tab:ablation_dn}, removing the ground-truth (GT) mask guidance causes the most significant mIoU drop (from 13.27\% to 12.95\%), highlighting its critical role in stabilizing the query-prototype association. The exclusion of semantic and feature noise also degrades performance, validating our full DN design.

\begin{table}[ht]
    \centering
    \caption{Analysis of Prototype Ratio $\rho$ in SPOT-CA. We investigate the impact of varying the prototype selection ratio $\rho$ on model accuracy and inference speed. The best trade-off is achieved at $\rho=0.08$.}
    \label{tab:ablation_rho}
    \resizebox{0.7\linewidth}{!}{%
    \begin{tabular}{c|cc}
    \toprule
    Ratio $\rho$ & mIoU (\%) $\uparrow$ & Overall Latency (ms) $\downarrow$ \\
    \midrule
    0.02 & 13.06 & 149 \\    
    0.04 & 13.15 & 152 \\
    0.06 & 13.22 & 156 \\
    \rowcolor{gray!20}
    0.08 & \textbf{13.27} & 164 \\
    0.10 & 13.21 & 175 \\
    0.12 & 13.19 & 193 \\
    \bottomrule
    \end{tabular}%
    }
\end{table}

\begin{table}[ht]
    \centering
    \caption{Analysis of Denoising (DN) Training Components. We dissect the contributions of different components within our DN training strategy to validate our design choices.}
    \label{tab:ablation_dn}
    \resizebox{0.6\linewidth}{!}{%
    \begin{tabular}{l|c}
    \toprule
    Configuration & mIoU (\%) $\uparrow$ \\
    \midrule
    Full DN Training (Ours) & \textbf{13.27} \\
    \quad w/o Semantic Noise & 13.04 \\
    \quad w/o Feature Noise & 13.11 \\
    \quad w/o GT Mask Guidance & 12.95 \\
    \bottomrule
    \end{tabular}%
    }
\end{table}

\section{CONCLUSIONS}

In this paper, we challenge the paradigm of global query-voxel interaction for the decoder stage in sparse 3D occupancy prediction. We posit that efficiency and accuracy are best achieved not by attending to all features, but by intelligently selecting a few.
Our framework, SPOT-Occ, embodies this principle through a two-stage process: guided feature selection and focused aggregation. Its Sparse Prototype Selection mechanism dynamically creates a compact, salient set of voxel features for each query, an inherently unstable process we stabilize with a dedicated denoising training paradigm.
The efficacy of our approach is validated on both the nuScenes-Occupancy and SemanticKITTI datasets; it not only achieves a higher mIoU but also reduces inference latency by 57.6\% on nuScenes, showcasing a remarkable improvement in efficiency. This work demonstrates that overcoming the efficiency bottleneck in 3D decoders lies not in refining dense query-voxel interactions, but in replacing them with smarter, sparse prototype-guided selection mechanisms.


\newpage
\bibliographystyle{IEEEtran}
\bibliography{ref}

\end{document}